\documentclass[11pt,a4paper]{researcharticle}
\usepackage{array,ragged2e,booktabs,makecell,longtable}
\usepackage{adjustbox,xurl,placeins}
\usepackage{natbib}
\setcitestyle{authoryear,round,aysep={,},citesep={;}}
\usepackage{fontawesome5}
\newcommand{\hficon}{\raisebox{-0.15em}{\includegraphics[height=1.15em]{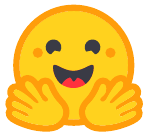}}}
\DeclareUnicodeCharacter{0394}{\ensuremath{\Delta}}
\DeclareUnicodeCharacter{03C1}{\ensuremath{\rho}}
\DeclareUnicodeCharacter{2081}{\textsubscript{1}}
\DeclareUnicodeCharacter{207B}{\textsuperscript{-}}
\DeclareUnicodeCharacter{2076}{\textsuperscript{6}}

\hypersetup{colorlinks=true,allcolors=darkblue}
\setheadertext{Tool-Augmented On-Policy Distillation}
\renewcommand{\today}{}

\title{Tool-Augmented On-Policy Distillation for LLM Domain Adaptation in Sequence-Based Omics Tasks}
\author{%
Jie Ying$^{1}$, Zhefan Wang$^{1}$, Zihong Chen$^{1}$, Zhengqing Li$^{2}$, Jinzhe Li$^{1}$, Gang Li$^{1}$, Jian Liu$^{1}$, Fang Hu$^{3}$, Tao Luo$^{2}$, Zhonghang Yuan$^{1}$, Wanli Ouyang$^{1,4}$, Stan Z. Li$^{5}$, Fan Yang$^{2}$, and Nanqing Dong$^{1,3,\dagger}$\\
$^{1}$Shanghai Artificial Intelligence Laboratory\quad
$^{2}$Yazhouwan National Laboratory\quad
$^{3}$Shanghai Innovation Institute\quad
\mbox{$^{4}$The Chinese University of Hong Kong}\quad
$^{5}$Westlake University
\vskip -1mm
\textbf{$^{\dagger}$Corresponding Author.}
\vskip -1mm
\faGithub~{\fontfamily{ppl}\selectfont\textbf{Benchmark:}}~\url{https://github.com/SeedScientist/OmicsBench}\\
\vskip 1mm
\hficon~{\fontfamily{ppl}\selectfont\textbf{Dataset:}}~\url{https://huggingface.co/datasets/yj12869741/TA-OPD-10K}\\
\vskip 1mm
\hficon~{\fontfamily{ppl}\selectfont\textbf{Model:}}~\url{https://huggingface.co/yj12869741/TA-OPD-27B}
}
\begin{abstract}
Multi-omics sequences contain complex biological patterns, yet deciphering the underlying mechanisms for automated scientific discovery remains a fundamental challenge. As large language models (LLMs) are increasingly deployed to interpret these sequences, it becomes critical to evaluate not just their predictions, but their capacity for scientific reasoning. However, existing benchmarks for multi-omics sequence tasks predominantly rely on black-box classification and regression metrics, neglecting whether models truly grasp the biological evidence within the multi-omics data. To address this gap, we introduce \textbf{OmicsBench}, the first reasoning benchmark for multi-omics sequences, comprising 1,160 expert-validated questions across six biologically coherent tasks spanning DNA regulation, RNA processing, and protein function. Unlike traditional benchmarks, OmicsBench requires LLMs to provide traceable evidence chains, which are strictly evaluated using instance-specific rubrics developed with domain experts. Our evaluation of 17 LLMs reveals room for improvement in multi-omics reasoning. We observe an inverse relationship: while scientific LLMs outperform general-purpose LLMs in classification accuracy on biological sequence understanding tasks, they fail to provide valid evidence to support their predictions. One plausible interpretation is shortcut learning: specialized models may rely on statistical patterns rather than the biological mechanisms needed for scientific discovery. Motivated by this finding, we further introduce \textbf{tool-augmented on-policy distillation (TA-OPD)}, a post-training method designed to align sequence prediction with evidence-grounded biological reasoning. Across five Qwen3.5 models spanning 0.8B to 27B parameters, TA-OPD consistently strengthens biological evidence grounding while improving predictive performance on most tasks. These gains persist across model scales, indicating that stronger sequence reasoning does not arise solely from increased model capacity, but can be improved through evidence-aware training. Together, OmicsBench and TA-OPD provide both a rigorous framework for diagnosing reasoning failures in multi-omics LLMs and a practical path toward models whose predictions are better grounded in biologically meaningful evidence.
\end{abstract}
\begin{document}
\maketitle
\section{Introduction}

Multi-omics sequences, including DNA, RNA, and proteins, represent an evolutionary heritage encoding billions of years of biological innovation~\citep{crick1970,ihgsc2004}. Deciphering the information embedded in these sequences is fundamental to understanding disease mechanisms~\citep{hasin2017}, designing novel therapeutics~\citep{jiang2025}, and enabling automated scientific discovery in the life sciences~\citep{swanson2025}.

Recent advances in biological foundation models have created powerful tools for multi-omics analysis. For example, Evo~\citep{nguyen2024} deciphers long-range dependencies within genomic sequences, RNA-FM~\citep{chen2022} captures the complex processing logic of transcriptomic sequences, and ESM~\citep{lin2023} learns functional representations of proteomic evolution. Together, these foundation models enable unprecedented capabilities in sequence-to-function prediction.

Additionally, the emergence of large language models (LLMs) offers a complementary opportunity to advance scientific discovery through scientific data analysis and automated scientific workflows \citep{boiko2023,ghareeb2026}. Recent studies illustrate this potential through biological hypothesis generation \citep{penades2025}, nanobody design \citep{swanson2025}, and gene-set interpretation grounded in domain databases \citep{wang2025}. In multi-omics research, LLMs have shown potential to orchestrate specialized foundation models, interpret multi-omics findings, and generate scientific hypotheses \citep{huang2026}, with the broader goal of accelerating scientific discovery. Yet realizing this vision requires LLMs that can assess the validity of biological predictions, understand the underlying reasoning processes, and provide scientifically grounded explanations.

While LLMs have advanced in multi-omics domains~\citep{dealmeida2025,xia2025}, we highlight two key limitations in the current paradigm. First, mainstream training typically relies on large-scale annotated datasets to map high-dimensional sequences to brief textual descriptions. This ``long sequence to short text'' setup encourages models to memorize direct mappings instead of learning the intermediate biological mechanisms or discerning the biological significance of sequence variations. Without chain-of-thought supervision during training, these models degenerate into sophisticated pattern matchers rather than scientific reasoners. Second, existing benchmarks predominantly evaluate outcome-oriented metrics (\emph{e.g.}, classification accuracy for sequence labeling and regression error for property prediction), without requiring models to articulate the biological evidence supporting their answers.

To the best of our knowledge, no existing benchmark provides a systematic framework for evaluating LLM reasoning capabilities across multi-omics sequence analysis tasks. Although prior benchmarks incorporate explicit multi-omics sequences within the input prompts, evaluation is still largely outcome-centric, focusing on predictive performance in classification, regression, or cloze-style completion~\citep{jin2024,laurent2024,shang2025}. Furthermore, the lack of standardized methodologies for generating and evaluating long-form biological reasoning traces across diverse multi-omics sequences has hindered process-oriented evaluation. Here, prediction refers to assigning a biological label or estimating a biological property from an input sequence, such as identifying a promoter or predicting enzyme function. Biological reasoning refers to explaining how relevant biological evidence, such as sequence motifs, homology, or functional annotations, supports that prediction.

To address these challenges, we introduce \textbf{OmicsBench}, the first benchmark for evaluating LLM reasoning capabilities in multi-omics sequence analysis. OmicsBench comprises 1,160 expert-validated questions across six biologically coherent tasks that domain experts would consider when analyzing multi-omics data. These tasks are organized along the sequential logic of multi-omics information processing: DNA regulation (identifying epigenetic marks, promoter regions, and transcription factor binding sites), RNA processing (characterizing modifications and non-coding RNAs), and protein function (annotating enzyme functions). To generate high-quality reasoning traces at scale, we design a multi-agent synthesis framework that deploys tool-augmented bio-agents to query biological databases, perform sequence alignments, and retrieve literature evidence to automatically curate traceable reasoning chains. All questions and solutions undergo two-tier validation via machine-based checks and expert reviews. This approach suggests a blueprint for constructing reasoning-oriented scientific benchmarks at scale. To evaluate long-form reasoning, we introduce Rubric Recall (RR) within an expert-in-the-loop evaluation protocol that builds on rubric-based assessment \citep{arora2025}. First, domain experts define task-level evaluation dimensions (\emph{e.g.}, motif identification, homology evidence, and mechanistic explanation). Then, for each question, we decompose the curated reasoning into atomic evidence units (\emph{e.g.}, TATA box at position \(- 30\) or BLAST E-value \(\text{<}10^{- 10}\)). Finally, we cluster these observed facts into instance-specific rubrics, expecting LLMs to recall as much biological evidence as possible to simulate expert thinking and problem-solving. We evaluate 17 leading LLMs on OmicsBench, spanning proprietary, open-source, and scientific LLMs. Our experiments highlight a critical phenomenon that we term ``reasoning degeneration''. While scientific LLMs excel at sequence classification, they fail significantly in articulating the supporting biological evidence. We hypothesize that this disparity may stem from shortcut learning~\citep{geirhos2020}: models may capture simplistic patterns or spurious correlations within the sequences, potentially attaining high accuracy without fully grounding their predictions in the biological evidence and reasoning required for scientific discovery. This gap between prediction and biological reasoning raises a natural question: can LLMs be trained to better connect their predictions to the underlying biological evidence?

Leading general-purpose LLMs, including GPT and Claude models, still have difficulty with sequence-based omics tasks. These models learn extensive biomedical knowledge during pretraining \citep{singhal2023}, but do not always apply it effectively to a given sequence and question. They may give a correct label without explaining its biological basis, or offer a plausible explanation that does not match the relevant evidence \citep{arora2026}. In contrast, biomedical AI agents can use specialized tools and databases to analyze biological sequences. Biomni, for example, combines LLMs with BLAST for sequence similarity searches, biological databases, Python code execution, and AI-based prediction models to answer biomedical questions and support scientific discovery \citep{huang2026}. The evidence they provide for DNA, RNA, and protein questions can be checked against tool outputs or database records. This motivates us to investigate whether LLMs can learn from these agents to improve their predictions and biological explanations without using tools at inference.

To this end, we use knowledge distillation to help LLMs learn from the agents' answers and supporting evidence. On-policy distillation (OPD) is suited to this goal because the student receives teacher guidance at each token of its own response, rather than only on the final answer \citep{agarwal2024}. This lets us train both biological explanations and predictions. On-policy self-distillation (OPSD) uses the same starting model for teacher and student, but gives only the teacher access to privileged information, such as ground-truth answers \citep{zhao2026}. The extra context can therefore help the teacher provide more informative guidance without requiring a larger model. Building on OPSD, we introduce \textbf{tool-augmented on-policy distillation (TA-OPD)}, a post-training method that uses the agents' answers and supporting biological evidence as privileged information provided only to the teacher during training.

We evaluated TA-OPD across five Qwen3.5 model sizes, from 0.8B to 27B parameters, to test whether prediction and biological explanations could improve together. We selected Qwen3.5 because Qwen is one of the most popular open-weight model families, is widely used in academic research, and offers a broad range of model sizes. We refer to Qwen's officially released instruction-tuned models before TA-OPD as unadapted models. Compared with the unadapted models, average RR across the six tasks increased at all five sizes, and mean predictive scores increased on all six tasks at 9B and 27B. The adapted 27B model also exceeded the unadapted Qwen3.5-397B reference in mean predictive scores on five of six tasks and in average RR. These results show that predictive performance and biological evidence grounding can be improved together through TA-OPD. We release the 10,069-record TA-OPD-10K training dataset and the TA-OPD-27B model weights for further training and to help reproduce our results.

Herein, we develop OmicsBench by combining agentic synthesis of expert-validated reasoning traces with rubric-based assessment of biological evidence coverage beyond surface-level text overlap. Using this benchmark, we evaluate 17 leading LLMs and identify critical gaps between sequence prediction and biological reasoning. To address these gaps, we explore TA-OPD as a post-training method for improving both predictive performance and reasoning quality, and release the TA-OPD-10K training dataset and TA-OPD-27B model weights to support further research. Together, these efforts provide a roadmap for training more capable LLMs for scientific reasoning across DNA regulation, RNA processing, and protein function.

\FloatBarrier
\section{Results}

\subsection{Experimental setup}

We compared the performance of 17 LLMs on OmicsBench, including 5 proprietary models, 6 open-source models, and 6 scientific LLMs. This comprehensive comparison not only highlights the relative strengths and limitations of each group but also provides insights for future research directions and applications. We use the following task labels: epigenetic mark prediction (EMP), promoter detection (Prom), transcription factor binding site prediction (TFBS), RNA modification prediction (Mod), non-coding RNA classification (ncRNA), and enzyme function prediction (EC). Predictive metrics include the Matthews correlation coefficient (MCC) and the area under the receiver operating characteristic curve (AUC).

\begin{table}[!htbp]
\centering
\caption{Comprehensive evaluation on OmicsBench. Left: Conventional prediction metrics assess the predictive performance. Right: Rubric Recall (RR, \%) evaluates the reasoning quality. Average rank (Avg Rank, left) and average RR (Avg Recall, right) summarize performance across diverse tasks. Avg Rank is used as conventional metrics differ across tasks and cannot be directly averaged. The columns delineate six task abbreviations: EMP (Epigenetic Mark Prediction), Prom (Promoter Detection), TFBS (Transcription Factor Binding Site Prediction), Mod (RNA Modification Prediction), ncRNA (Non-coding RNA Classification), and EC (Enzyme Function Prediction). MCC denotes Matthews correlation coefficient; AUC, area under the receiver operating characteristic curve; Acc, accuracy; and Fmax, maximum F1-score. `--' indicates unsupported tasks. Top-3 performers per column are highlighted in red: \colorbox{red!20}{1st}, \colorbox{red!10}{2nd}, and \colorbox{red!5}{3rd}.}
\label{tab:1}
\fontsize{8}{10}\selectfont
\setlength{\tabcolsep}{3pt}
\renewcommand{\arraystretch}{1.17}
\begin{adjustbox}{max width=\linewidth}
\begin{tabular}{@{}lcccccccccccccc@{}}
\toprule
\textbf{Model} & \makecell[l]{\textbf{EMP}\\
\textbf{(MCC)}} & \makecell[l]{\textbf{Prom}\\
\textbf{(MCC)}} & \makecell[l]{\textbf{TFBS}\\
\textbf{(MCC)}} & \makecell[l]{\textbf{Mod}\\
\textbf{(AUC)}} & \makecell[l]{\textbf{ncRNA}\\
\textbf{(Acc)}} & \makecell[l]{\textbf{EC}\\
\textbf{(Fmax)}} & \makecell[l]{\textbf{Avg}\\
\textbf{Rank \ensuremath{\downarrow}}} & \makecell[l]{\textbf{EMP}\\
\textbf{RR (\%)}} & \makecell[l]{\textbf{Prom}\\
\textbf{RR (\%)}} & \makecell[l]{\textbf{TFBS}\\
\textbf{RR (\%)}} & \makecell[l]{\textbf{Mod}\\
\textbf{RR (\%)}} & \makecell[l]{\textbf{ncRNA}\\
\textbf{RR (\%)}} & \makecell[l]{\textbf{EC}\\
\textbf{RR (\%)}} & \makecell[l]{\textbf{Avg}\\
\textbf{Recall \ensuremath{\uparrow}}}\\
\midrule
\addlinespace[3pt]
\multicolumn{15}{l}{\emph{\textbf{Proprietary LLMs}}}\\
Claude-Sonnet-4.5 & 3.56 & 28.58 & 12.50 & 53.42 & 12.56 & 8.04 & 7.67 & \cellcolor{red!20}23.27 & 25.60 & 12.88 & \cellcolor{red!5}20.63 & 9.49 & 4.77 & 16.11\\
Gemini-3-Pro & -10.22 & 16.61 & 2.32 & 51.57 & \cellcolor{red!10}24.19 & \cellcolor{red!5}24.68 & 9.67 & 4.74 & 22.21 & \cellcolor{red!5}24.96 & 8.17 & \cellcolor{red!10}24.31 & \cellcolor{red!10}12.60 & \cellcolor{red!5}16.17\\
GPT-5.2 & 14.38 & 15.97 & -0.81 & 53.17 & 3.72 & 8.62 & 11.67 & 4.70 & 20.00 & 21.77 & 7.26 & 8.74 & 4.07 & 11.09\\
Grok-4 & -3.50 & 30.67 & 2.09 & 50.28 & 13.95 & 23.39 & 9.33 & \cellcolor{red!5}10.79 & \cellcolor{red!20}30.16 & \cellcolor{red!20}27.46 & 10.81 & \cellcolor{red!20}34.04 & \cellcolor{red!20}13.34 & \cellcolor{red!20}21.10\\
Qwen3-Max & 8.36 & 21.35 & -0.98 & 53.10 & \cellcolor{red!5}19.07 & 10.91 & 9.00 & 6.30 & \cellcolor{red!10}28.89 & \cellcolor{red!10}25.39 & \cellcolor{red!20}29.95 & 16.07 & 4.55 & \cellcolor{red!10}18.53\\
\addlinespace[3pt]
\multicolumn{15}{l}{\emph{\textbf{Open-Source LLMs}}}\\
DeepSeek-V3.2 & 1.28 & 23.78 & 8.91 & 52.25 & 9.77 & 11.68 & 8.67 & \cellcolor{red!10}12.27 & 25.79 & 20.82 & 5.86 & 6.65 & 4.31 & 12.62\\
GLM-4.7 & -5.29 & 12.16 & 6.96 & 50.90 & 7.44 & 8.94 & 12.50 & 4.44 & 23.13 & 12.43 & 13.53 & 7.81 & 3.35 & 10.78\\
GPT-OSS-120B & \cellcolor{red!5}20.96 & 18.98 & 0.17 & 52.48 & 8.84 & 11.46 & 9.17 & 4.74 & 22.37 & 24.17 & 19.97 & 18.52 & 3.98 & 15.63\\
Kimi-K2 & 9.65 & 0.94 & 4.85 & 47.24 & 4.65 & 16.58 & 12.00 & 5.29 & 16.23 & 9.23 & 12.95 & 2.40 & 9.03 & 9.19\\
Llama-4-Maverick & 14.72 & 17.53 & 4.89 & \cellcolor{red!5}53.50 & 6.98 & 10.42 & 8.83 & 2.16 & 11.36 & 3.81 & 2.23 & 2.20 & 2.67 & 4.07\\
Qwen3-235B & 1.05 & 2.97 & 0.68 & \cellcolor{red!10}54.12 & 5.12 & 7.82 & 12.17 & 9.56 & \cellcolor{red!5}28.19 & 22.08 & \cellcolor{red!10}25.33 & 7.99 & 3.40 & 16.09\\
\addlinespace[3pt]
\multicolumn{15}{l}{\emph{\textbf{Scientific LLMs}}}\\
ChatMultiOmics & 10.44 & 20.66 & \cellcolor{red!5}21.83 & \cellcolor{red!20}59.72 & \cellcolor{red!20}83.26 & 23.12 & \cellcolor{red!5}4.33 & 1.27 & 4.07 & 1.59 & 0.00 & \cellcolor{red!5}20.80 & 2.65 & 5.06\\
ChatNT & \cellcolor{red!20}86.86 & \cellcolor{red!10}40.76 & 12.46 & -- & -- & -- & \cellcolor{red!20}3.00 & 0.00 & 0.00 & 0.25 & -- & -- & -- & 0.08\\
Intern-S1 & -0.11 & 19.24 & 21.81 & 51.36 & 11.63 & 13.56 & 9.00 & 2.33 & 20.45 & 12.39 & 12.05 & 2.44 & 3.67 & 8.89\\
Intern-S1-Pro & 20.61 & \cellcolor{red!20}47.93 & \cellcolor{red!10}43.95 & 52.27 & 17.67 & \cellcolor{red!10}39.53 & \cellcolor{red!10}3.50 & 0.00 & 0.23 & 0.00 & 0.00 & 3.99 & \cellcolor{red!5}10.23 & 2.41\\
NatureLM & -6.66 & \cellcolor{red!5}32.41 & 10.87 & 52.04 & 0.00 & 14.76 & 9.83 & 0.00 & 0.00 & 0.00 & 0.00 & 0.00 & 1.35 & 0.23\\
SciReasoner & \cellcolor{red!10}29.67 & 31.29 & \cellcolor{red!20}44.37 & 50.52 & 6.51 & \cellcolor{red!20}77.41 & 5.67 & 3.17 & 2.00 & 0.00 & 0.99 & 1.05 & 1.11 & 1.39\\
\bottomrule
\end{tabular}
\end{adjustbox}
\end{table}

\subsection{Evaluation of prediction and reasoning}

Table~\ref{tab:1} jointly reports conventional task metrics (predictive performance) and RR (\%) (reasoning quality), revealing a systematic trade-off between label-level accuracy and evidence coverage.

\textbf{Predictive performance.} Scientific LLMs are often the strongest on \emph{conventional metrics}. In fact, the best single-task conventional score in each column is achieved by scientific LLMs (\emph{e.g.}, ChatNT on EMP; Intern-S1-Pro on Prom; SciReasoner on TFBS and EC; ChatMultiOmics on Mod and ncRNA). This suggests that domain specialization can yield highly competitive label predictions on specific omics tasks. Consistently, scientific LLMs also attain the best average rank (Avg Rank) in Table~\ref{tab:1} (\emph{e.g.}, ChatMultiOmics: 4.33; Intern-S1-Pro: 3.50; ChatNT: 3.00 on the subset of DNA tasks it reports), while the best Avg Rank among proprietary general LLMs is Claude-Sonnet-4.5 (7.67), and the best among open-source general LLMs is DeepSeek-V3.2 (8.67).

\textbf{Reasoning quality.} In contrast, general-purpose LLMs dominate \emph{RR} in most tasks. The highest Avg Recall is achieved by Grok-4 (21.10\%), followed by Qwen3-Max (18.53\%) and Gemini-3-Pro (16.17\%). At the task level, general-purpose models attain the top RR on all six tasks. This indicates that explicit evidence recovery remains limited even when prediction is strong. Within open-source general LLMs, Qwen3-235B attains the best Avg Recall (16.09\%), while within scientific LLMs, Intern-S1 attains the best Avg Recall (8.89\%).

Overall, Table~\ref{tab:1} suggests that current scientific LLMs are often strong at producing correct labels on specific tasks, whereas general-purpose LLMs tend to provide better rubric-aligned evidence coverage. This motivates evaluating LLMs on multi-omics tasks with both accuracy and RR, since high predictive scores alone do not guarantee transparent, evidence-grounded explanations.

\subsection{Empirical analysis}

\begin{figure}[!htb]
\centering
\includegraphics[width=.60\linewidth]{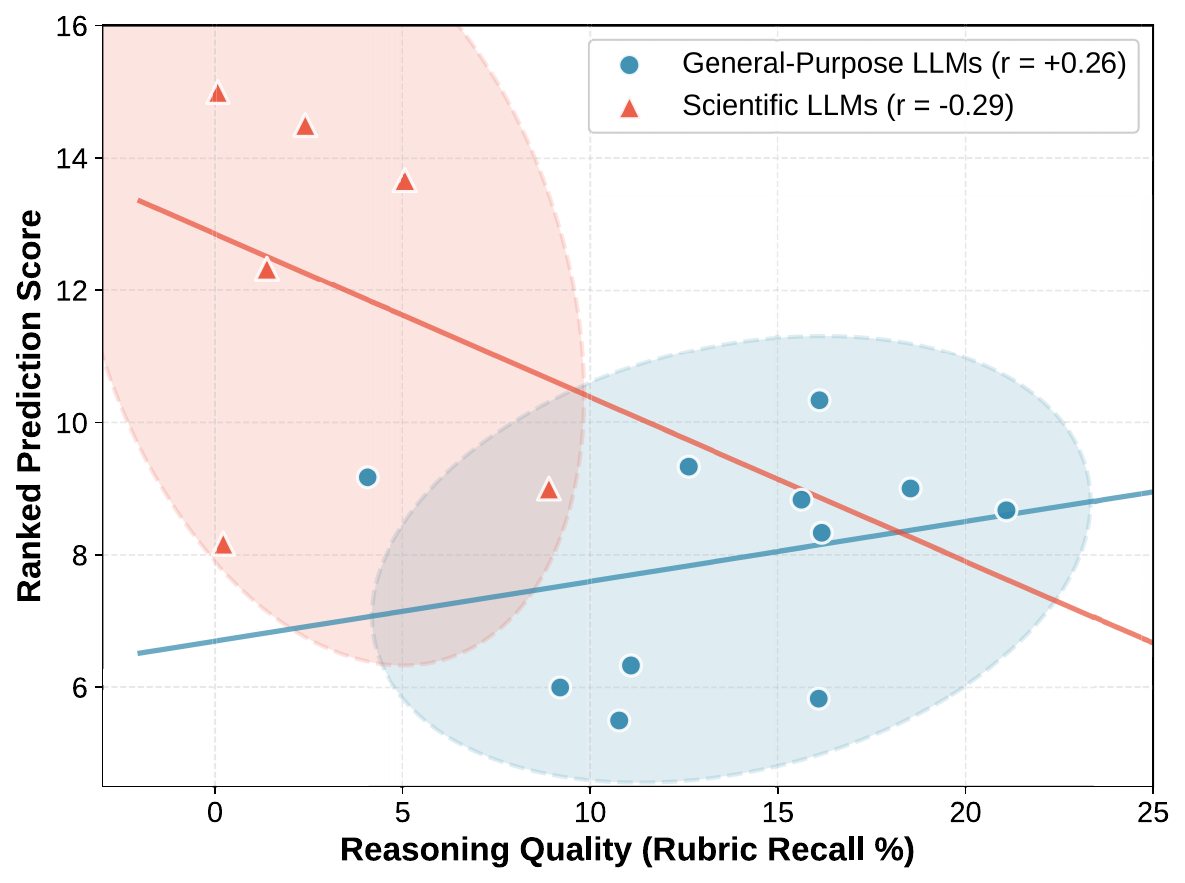}
\caption{Divergence between reasoning quality and prediction performance. The scatter plot compares general-purpose LLMs (circles) and scientific LLMs (triangles). (1) Axes: The \(x\)-axis represents Rubric Recall, while the \(y\)-axis denotes the Ranked Prediction Score (range: 1--17), calculated as \(N + 1 - \text{Avg\ Rank}\) (where \(N = 17\)). Higher scores indicate better performance; the score is rank-based rather than a raw MCC/AUC value. (2) Statistics: Solid lines represent linear regression fits (Ordinary Least Squares). Shaded regions indicate 95\% confidence ellipses (spanning 2 standard deviations from the mean). \(r\) values denote Pearson correlation coefficients. (3) Observation: While general LLMs show a positive correlation (\(r = 0.26\)), scientific LLMs exhibit a negative trend (\(r = - 0.29\)). Notably, scientific LLMs achieve prediction scores approximately 1.5\(\times\) those of general LLMs (12.11 vs. 7.94), but their reasoning quality is only about 0.2\(\times\) that of general LLMs (3.01\% vs. 13.76\%), suggesting that some scientific LLMs achieve high prediction accuracy without robust reasoning capabilities for multi-omics sequences}
\label{fig:1}
\end{figure}

\textbf{Phenomenon of reasoning degeneration.} From first principles, one might expect reasoning quality and predictive performance to be strongly aligned: models that truly ``reason'' with biological evidence should more often reach correct conclusions and cite the supporting cues. However, across all (model, task) pairs, we observe a statistically significant divergence in Pearson correlation across different model families. As shown in Figure~\ref{fig:1}, within proprietary and open-source general models, the correlation is moderate (\(r = 0.26\)), whereas within scientific LLMs it becomes negative (\(r = - 0.29\)). A concrete example is EMP, where ChatNT achieves 86.86\% MCC but 0\% RR under the same prompts that request biological reasoning. This pattern suggests a form of ``reasoning degeneration'': some models can output high-confidence labels without explicitly recovering rubric-aligned evidence. A plausible explanation is shortcut behavior, where models rely on surface patterns sufficient for label prediction but do not externalize the biological evidence required by the rubric. Consistent with this, scientific LLMs produce responses averaging only 47 tokens, compared to 857 tokens for general LLMs, suggesting short-form label generation rather than evidence-based reasoning. The results provide further evidence for the shortcut learning hypothesis: scientific LLMs, having been fine-tuned primarily on sequence classification objectives with short-form labels, tend to (1) produce significantly shorter outputs than general LLMs; (2) bypass detailed biological reasoning even when explicitly prompted; and (3) achieve high prediction accuracy through pattern matching rather than reasoning grounded in biological principles.

\textbf{Imbalance of multi-omics capabilities.} As shown in Table~\ref{tab:1}, multi-omics capabilities are uneven across modalities, characterized by substantial performance variance and the absence of a universally dominant model. On the DNA side, RR attains comparatively high peaks (\emph{e.g.}, 30.16\% on Prom and 27.46\% on TFBS), and on Mod it reaches 29.95\%. In contrast, the protein task (EC) exhibits much lower RR across models, with the highest value in our table being 13.34\%. This gap suggests that, under our rubric, recovering and articulating protein-function evidence is more challenging than citing motif-level or regulatory-sequence evidence. We also observe pronounced unevenness in model strengths across tasks: no single model appears in the top-3 across all six RR columns. For instance, Grok-4 ranks among the best on Prom and ncRNA RR but is less competitive on Mod RR, while Claude-Sonnet-4.5 is strong on EMP and Mod RR yet weaker on TFBS RR. Among scientific LLMs, ChatMultiOmics achieves a strong ncRNA RR (20.80\%), yet remains near-zero on several other RR columns. A similar lack of a universally best model is observed for conventional metrics: top-3 ranks also vary by task, indicating that strong predictive performance does not consistently transfer across modalities either. Overall, these patterns indicate that future work should focus on unified training strategies to integrate multi-omics capabilities, rather than relying on monolithic approaches.

\textbf{Qualitative error analysis.} To interpret the failures of LLMs on multi-omics tasks, we conduct an expert case study and highlight the three most critical failure modes. (1) Deterministic Amplification of Weak Signals: Models frequently treat statistical correlations (e.g., localized GC-rich regions in DNA or short sequence motifs in RNA) as definitive functional proof, ignoring the need for combinatorial evidence. (2) Hallucination of Biological States: Models often assert specific biological states (e.g., ``open chromatin'' or ``methylation'') without verifiable evidence in the sequence. (3) Fictitious Tool Results: In protein tasks, models often hallucinate the execution of external tools (e.g., ``BLAST shows high homology'') to justify predictions, despite having no access to such tools. These patterns are consistent with a shortcut-learning account, where high predictive accuracy may reflect memorization of textual surface patterns and stylistic mimicry rather than evidence-grounded reasoning.

\subsection{Predictive performance after TA-OPD}

Compared with their respective unadapted models, TA-OPD models achieved higher mean predictive scores on four of six tasks at 0.8B, five tasks at 2B and 4B, and all six tasks at 9B and 27B (Table~\ref{tab:2}). Larger models improved on more tasks in these experiments, although the gains varied across tasks. We also evaluated the unadapted Qwen3.5-397B model as an external reference using the same inference and scoring protocol. The TA-OPD 9B and 27B models each achieved higher mean predictive scores than this larger model on five tasks, with EC as the exception. At the task level, TA-OPD showed clear gains in EMP, Prom, TFBS and ncRNA, particularly at 27B. For Prom, MCC increased by 15.94, 20.50, 34.97 and 33.13 percentage points at 2B, 4B, 9B and 27B, respectively. TFBS MCC was higher at four of five scales, except at 2B. EMP gains varied across model sizes: the adapted 0.8B model scored below its unadapted counterpart, whereas gains reached +28.89 percentage points at 9B and +34.04 at 27B. ncRNA accuracy increased at all five model sizes, with the largest gain at 27B (from 5.12\% to 19.22\%). Changes in Mod AUC were small, while EC scores remained low.

\begin{table}[!htbp]
\centering
\caption{Predictive performance on OmicsBench}
\label{tab:2}
\fontsize{8}{10}\selectfont
\setlength{\tabcolsep}{3pt}
\renewcommand{\arraystretch}{1.17}
\begin{adjustbox}{max width=\linewidth}
\begin{tabular}{l|ccccccc}
\toprule
\textbf{Method} & \makecell[c]{\textbf{EMP}\\
\textbf{(MCC)}} & \makecell[c]{\textbf{Prom}\\
\textbf{(MCC)}} & \makecell[c]{\textbf{TFBS}\\
\textbf{(MCC)}} & \makecell[c]{\textbf{Mod}\\
\textbf{(AUC)}} & \makecell[c]{\textbf{ncRNA}\\
\textbf{(Acc)}} & \makecell[c]{\textbf{EC}\\
\textbf{(Fmax)}} & \textbf{Tasks improved}\\
\midrule
\multicolumn{8}{l}{\emph{Qwen3.5-0.8B}}\\
\quad Unadapted & \textbf{3.28 ± 5.07} & 5.14 ± 3.24 & \ensuremath{-}3.82 ± 3.65 & 52.49 ± 2.33 & 6.67 ± 0.97 & \textbf{1.06 ± 0.02} & NA\\
\rowcolor{gray!20}\quad + TA-OPD & 2.31 ± 2.95 & \textbf{5.20 ± 3.15} & \textbf{2.96 ± 2.41} & \textbf{53.23 ± 1.82} & \textbf{7.60 ± 0.71} & 1.01 ± 0.01 & 4/6\\
\midrule
\multicolumn{8}{l}{\emph{Qwen3.5-2B}}\\
\quad Unadapted & 0.78 ± 2.81 & \ensuremath{-}0.67 ± 6.17 & \textbf{0.35 ± 2.90} & 48.87 ± 2.32 & 7.29 ± 0.97 & 0.99 ± 0.01 & NA\\
\rowcolor{gray!20}\quad + TA-OPD & \textbf{5.83 ± 2.74} & \textbf{15.27 ± 1.89} & \ensuremath{-}2.04 ± 5.09 & \textbf{48.92 ± 0.62} & \textbf{8.84 ± 0.80} & \textbf{1.71 ± 1.28} & 5/6\\
\midrule
\multicolumn{8}{l}{\emph{Qwen3.5-4B}}\\
\quad Unadapted & 0.35 ± 6.18 & 1.52 ± 1.71 & 3.67 ± 5.71 & \textbf{51.71 ± 1.91} & 8.84 ± 0.47 & 0.96 ± 0.03 & NA\\
\rowcolor{gray!20}\quad + TA-OPD & \textbf{8.52 ± 5.29} & \textbf{22.01 ± 3.39} & \textbf{9.00 ± 3.97} & 51.26 ± 1.10 & \textbf{11.78 ± 0.27} & \textbf{1.70 ± 0.70} & 5/6\\
\midrule
\multicolumn{8}{l}{\emph{Qwen3.5-9B}}\\
\quad Unadapted & \ensuremath{-}4.32 ± 4.45 & 2.96 ± 6.64 & 7.42 ± 3.85 & 51.61 ± 4.33 & 13.95 ± 1.23 & 1.25 ± 0.62 & NA\\
\rowcolor{gray!20}\quad + TA-OPD & \textbf{24.57 ± 6.20} & \textbf{37.93 ± 2.05} & \textbf{22.19 ± 0.91} & \textbf{51.84 ± 1.72} & \textbf{14.11 ± 1.17} & \textbf{1.81 ± 1.07} & 6/6\\
\midrule
\multicolumn{8}{l}{\emph{Qwen3.5-27B}}\\
\quad Unadapted & \ensuremath{-}5.69 ± 1.42 & 6.10 ± 3.64 & 16.29 ± 1.70 & 50.83 ± 2.19 & 5.12 ± 0.47 & 0.90 ± 0.02 & NA\\
\rowcolor{gray!20}\quad + TA-OPD & \textbf{28.35 ± 1.27} & \textbf{39.22 ± 0.25} & \textbf{23.18 ± 1.14} & \textbf{53.52 ± 2.19} & \textbf{19.22 ± 0.54} & \textbf{1.84 ± 0.70} & 6/6\\
\midrule
\multicolumn{8}{l}{\emph{External reference}}\\
\quad Qwen3.5-397B & 7.40 ± 3.33 & 17.32 ± 3.66 & 12.32 ± 2.14 & 49.43 ± 0.12 & 6.51 ± 0.93 & 6.99 ± 1.33 & NA\\
\bottomrule
\end{tabular}
\end{adjustbox}
\par\vspace{4pt}
\begin{minipage}{\linewidth}\footnotesize EMP (epigenetic mark prediction), Prom (promoter detection) and TFBS (transcription factor binding site prediction) are reported using the Matthews correlation coefficient (MCC). Mod (RNA modification prediction) is reported as the area under the receiver operating characteristic curve (AUC). ncRNA (non-coding RNA classification) is reported as accuracy, and EC (enzyme function prediction) as the maximum F1-score (Fmax). NA denotes not applicable. Values are mean ± standard deviation, summarized as described in Materials and methods. Unadapted denotes the official instruction-tuned model before TA-OPD. ``Tasks improved'' counts tasks with a higher TA-OPD mean than the corresponding unadapted model; bold identifies the higher mean in each Unadapted/TA-OPD comparison. All predictive metrics are multiplied by 100 for display.\end{minipage}
\end{table}

\subsection{Reasoning quality after TA-OPD}

\begin{figure}[!htb]
\centering
\includegraphics[width=.60\linewidth]{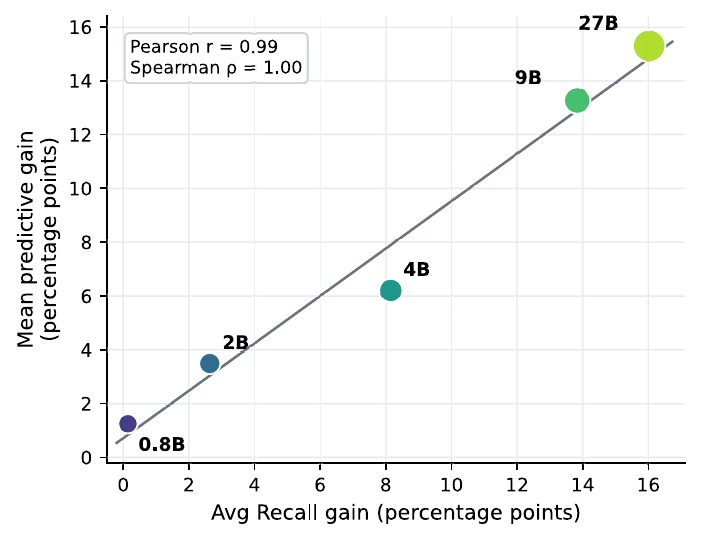}
\caption{Association between predictive and Rubric Recall gains across model sizes. Each point represents one model size; the horizontal axis shows the gain in Avg Recall over the corresponding unadapted model. The vertical axis is the arithmetic mean of the changes in predictive metrics across the six tasks, (ΔEMP + ΔProm + ΔTFBS + ΔMod + ΔncRNA + ΔEC)/6, combining MCC, AUC, accuracy and Fmax changes expressed in percentage points. Point size indicates parameter count, and the line is an ordinary least-squares fit. The correlations describe these five points (Pearson r = 0.99, Spearman ρ = 1.00) and do not establish a causal relationship between predictive improvement and evidence coverage}
\label{fig:2}
\end{figure}

We used RR to assess biological evidence in model responses using the OmicsBench rubric for each question. Avg Recall, which averages RR across the six tasks, increased at every model size, with gains ranging from 0.14 percentage points at 0.8B to 16.02 at 27B (Table~\ref{tab:3}). In the preceding OmicsBench evaluation, some scientific LLMs combined high predictive performance with low RR. After TA-OPD, larger gains in mean predictive scores were accompanied by larger gains in Avg Recall across the five model sizes (Figure~\ref{fig:2}). The 9B and 27B models reached Avg Recall values of 21.04\% and 25.66\%, respectively, compared with 11.64\% for the 397B reference. At 27B, EMP, Prom, TFBS and ncRNA showed clear gains in both prediction and RR. For ncRNA, the accuracy gain of 14.10 percentage points was accompanied by an increase in RR from 3.58\% to 12.48\%. Mod showed a different pattern: RR increased markedly at larger model sizes while predictive AUC changed little. EC scores remained low, and its RR decreased from 0.36\% to 0.16\% at 27B.

\begin{table}[!htbp]
\centering
\caption{Task-specific Rubric Recall on OmicsBench}
\label{tab:3}
\fontsize{8}{10}\selectfont
\setlength{\tabcolsep}{3pt}
\renewcommand{\arraystretch}{1.17}
\begin{adjustbox}{max width=\linewidth}
\begin{tabular}{l|ccccccc}
\toprule
\textbf{Method} & \makecell[c]{\textbf{EMP}\\
\textbf{RR (\%)}} & \makecell[c]{\textbf{Prom}\\
\textbf{RR (\%)}} & \makecell[c]{\textbf{TFBS}\\
\textbf{RR (\%)}} & \makecell[c]{\textbf{Mod}\\
\textbf{RR (\%)}} & \makecell[c]{\textbf{ncRNA}\\
\textbf{RR (\%)}} & \makecell[c]{\textbf{EC}\\
\textbf{RR (\%)}} & \makecell[c]{\textbf{Avg Recall}\\
\textbf{(\%)}}\\
\midrule
\multicolumn{8}{l}{\emph{Qwen3.5-0.8B}}\\
\quad Unadapted & 0.22 ± 0.10 & 0.31 ± 0.00 & 0.19 ± 0.00 & 0.00 ± 0.00 & 0.36 ± 0.35 & 0.00 ± 0.00 & 0.21 ± 0.02\\
\rowcolor{gray!20}\quad + TA-OPD & \textbf{0.38 ± 0.25} & \textbf{0.57 ± 0.09} & \textbf{0.25 ± 0.11} & \textbf{0.42 ± 0.18} & \textbf{0.43 ± 0.15} & 0.00 ± 0.00 & \textbf{0.35 ± 0.06}\\
\midrule
\multicolumn{8}{l}{\emph{Qwen3.5-2B}}\\
\quad Unadapted & 0.49 ± 0.43 & 1.77 ± 0.39 & 1.36 ± 0.43 & 0.21 ± 0.18 & 0.53 ± 0.25 & 0.10 ± 0.09 & 0.80 ± 0.23\\
\rowcolor{gray!20}\quad + TA-OPD & \textbf{4.12 ± 0.17} & \textbf{8.22 ± 0.65} & \textbf{6.18 ± 0.88} & \textbf{1.27 ± 0.00} & \textbf{1.00 ± 0.10} & \textbf{0.16 ± 0.00} & \textbf{3.43 ± 0.23}\\
\midrule
\multicolumn{8}{l}{\emph{Qwen3.5-4B}}\\
\quad Unadapted & 0.93 ± 0.58 & 8.68 ± 0.50 & 9.77 ± 1.88 & 3.72 ± 0.49 & 1.00 ± 0.20 & \textbf{0.16 ± 0.00} & 3.91 ± 0.27\\
\rowcolor{gray!20}\quad + TA-OPD & \textbf{10.21 ± 0.33} & \textbf{30.79 ± 0.74} & \textbf{18.37 ± 0.67} & \textbf{11.58 ± 1.77} & \textbf{2.32 ± 0.57} & 0.10 ± 0.18 & \textbf{12.07 ± 0.33}\\
\midrule
\multicolumn{8}{l}{\emph{Qwen3.5-9B}}\\
\quad Unadapted & 1.04 ± 0.25 & 15.65 ± 0.89 & 18.18 ± 0.81 & 1.70 ± 0.18 & 5.77 ± 0.43 & 0.16 ± 0.00 & 7.21 ± 0.24\\
\rowcolor{gray!20}\quad + TA-OPD & \textbf{24.83 ± 1.26} & \textbf{38.12 ± 0.48} & \textbf{33.27 ± 0.57} & \textbf{22.61 ± 1.10} & \textbf{6.57 ± 1.19} & 0.16 ± 0.00 & \textbf{21.04 ± 0.43}\\
\midrule
\multicolumn{8}{l}{\emph{Qwen3.5-27B}}\\
\quad Unadapted & 7.42 ± 0.16 & 19.40 ± 0.95 & 24.30 ± 0.81 & 2.12 ± 0.18 & 3.58 ± 0.36 & \textbf{0.36 ± 0.24} & 9.63 ± 0.36\\
\rowcolor{gray!20}\quad + TA-OPD & \textbf{35.69 ± 0.38} & \textbf{46.80 ± 0.41} & \textbf{36.61 ± 1.58} & \textbf{22.82 ± 1.44} & \textbf{12.48 ± 0.51} & 0.16 ± 0.16 & \textbf{25.66 ± 0.69}\\
\midrule
\multicolumn{8}{l}{\emph{External reference}}\\
\quad Qwen3.5-397B & 2.09 ± 0.41 & 22.78 ± 0.71 & 28.45 ± 0.88 & 1.17 ± 0.18 & 12.70 ± 0.40 & 2.81 ± 0.27 & 11.64 ± 0.15\\
\bottomrule
\end{tabular}
\end{adjustbox}
\par\vspace{4pt}
\begin{minipage}{\linewidth}\footnotesize Task abbreviations follow Table~\ref{tab:2}; Rubric Recall (RR) values are percentages, summarized as described in Materials and methods. Avg Recall gives equal weight to the six tasks. Bold denotes the higher mean in each Unadapted/TA-OPD comparison.\end{minipage}
\end{table}

\section{Discussion}

\subsection{Implications of OmicsBench}

In this study, we propose OmicsBench, the first benchmark dedicated to evaluating the reasoning capabilities of LLMs in multi-omics sequence analysis. The contributions of OmicsBench to the AI for Science community are twofold. First, it establishes a scalable blueprint for constructing reasoning-oriented scientific benchmarks. By introducing an agentic synthesis framework that orchestrates bio-agents with domain tools, we demonstrate how to overcome the scarcity of expert-annotated reasoning chains without compromising scientific rigor. This methodology can be transferred to other data-scarce scientific domains to bridge the gap between unstructured reasoning traces and structured evaluation. Second, OmicsBench provides a critical diagnostic for the current state of biological LLMs. Our evaluation reveals a systematic ``reasoning degeneration'' where scientific LLMs achieve high classification accuracy but fail to articulate the supporting biological evidence. This empirical finding challenges the current ``long sequence to short text'' training paradigm and encourages the community to seek solutions to the observed gap between predictive accuracy and mechanistic understanding. By shifting the evaluation focus from outcome prediction to evidence coverage, OmicsBench aims to serve as a practical foundation for developing the next generation of transparent and reasoning-capable bio-AI systems. These findings should nevertheless be interpreted within the scope of OmicsBench: RR measures expert-verifiable biological evidence rather than models' internal reasoning, predefined rubrics may omit valid but unanticipated evidence, and undisclosed training corpora prevent us from fully excluding benchmark overlap. Future evaluations could therefore complement rubric-based scoring with more flexible measures of biological reasoning.

\subsection{Implications of TA-OPD}

TA-OPD offers a way to transfer biomedical agents' ability to reason from evidence to LLMs through post-training. Across the five model sizes, larger mean predictive gains were accompanied by larger increases in Avg Recall. At 27B, EMP, Prom, TFBS and ncRNA showed clear gains in both prediction and RR without tool use during evaluation. This joint improvement addresses the gap between predicting a label and explaining the biological premises behind it. The training design of TA-OPD also allows a model accessed through a commercial application programming interface (API) to contribute to OPD without exposing its token probabilities. The API model guides the agent's tool use and produces answers with supporting evidence. These outputs serve as privileged information for a local teacher, which computes the token-level training signal. The student can therefore learn from both the API model's explanations and the biological evidence returned by domain tools.

Preparing agent-provided privileged information for training requires balancing evidence quality, sample coverage and annotation cost. Keeping only responses with correct final predictions helps avoid distilling incorrect answers, but it also favors questions the agent can already solve and can skew class proportions within each task. Stricter checks of every reasoning step would require more annotation and could exclude useful examples, whereas keeping all responses would expose the student to more incorrect reasoning. In this study, we chose answer-label consistency as the main filter and oversampled minority classes to improve their representation. The source of privileged information also matters. A teacher can appear more capable because it sees information unavailable to the student, leading the student to imitate outcomes without learning how to reproduce them. This problem has been described as the ``privilege illusion'' \citep{yu2026}. In TA-OPD, the teacher receives evidence obtained by analyzing the same sequence and question given to the student. Biological tool outputs and database records connect this evidence to the input sequence, giving the student guidance on the biological premises and analysis behind a prediction. This connection may reduce the risk of privilege illusion and help limit hallucinated biological explanations. We also observed a tendency toward longer responses during TA-OPD training. Previous work on OPD has linked excessive response-length inflation to repetition \citep{luo2026}. In sequence-based omics tasks, however, additional response length could be used to explain sequence features and their biological implications. Token-level guidance may help direct this extra text toward relevant evidence, preserving biological explanations and keeping LLMs from functioning only as classifiers or regressors. TA-OPD could also support continual learning as biological knowledge changes. Indeed, self-distillation has been shown to help models acquire new knowledge while retaining earlier capabilities \citep{shenfeld2026}. In TA-OPD, domain tools could supply updated privileged information for successive rounds of training, allowing new knowledge to be gradually incorporated into model weights. Thus, knowledge obtained for earlier questions could remain available after the privileged information is removed. Since biological tool use already includes online database searches, this suggests a broader extension to retrieval-augmented OPSD. The model could then continue learning as it interacts with external resources, receives external feedback, and retrieves new information from the internet. By supporting continued information acquisition and improvements to both teacher and student models, this iterative process could also contribute to recursive self-improvement \citep[RSI;][]{schmidhuber2007}.

\subsection{Limitations and future directions}

Although our results demonstrate the potential of TA-OPD to improve both sequence prediction and biological reasoning, our evaluation covers only six sequence tasks, and our post-training experiments are limited to models with up to 27B parameters. These limitations arise from both the substantial cost of constructing and validating evidence-grounded reasoning data and the computational resources required to extend post-training to models with hundreds of billions or trillions of parameters. In the future, more affordable data construction methods and computing hardware, together with more cost-effective training frameworks for LLMs, could help overcome these constraints, allowing researchers to explore the extension of TA-OPD to a broader range of tasks and larger model scales. An important future direction is to investigate how evidence-grounded prediction and reasoning can support scientific discovery, hypothesis generation, and experimental design through natural-language interaction under the supervision and guidance of domain experts. OmicsBench and TA-OPD provide a starting point for exploring this direction by evaluating and improving LLMs' ability to provide biologically grounded explanations for their sequence predictions.

\FloatBarrier
\section{Materials and methods}

This section introduces the design principles of OmicsBench, summarizes its six multi-omics tasks, and describes the benchmark construction and evaluation framework.

\subsection{Biologically grounded design}

\begin{figure}[!htbp]
\centering
\includegraphics[width=\linewidth]{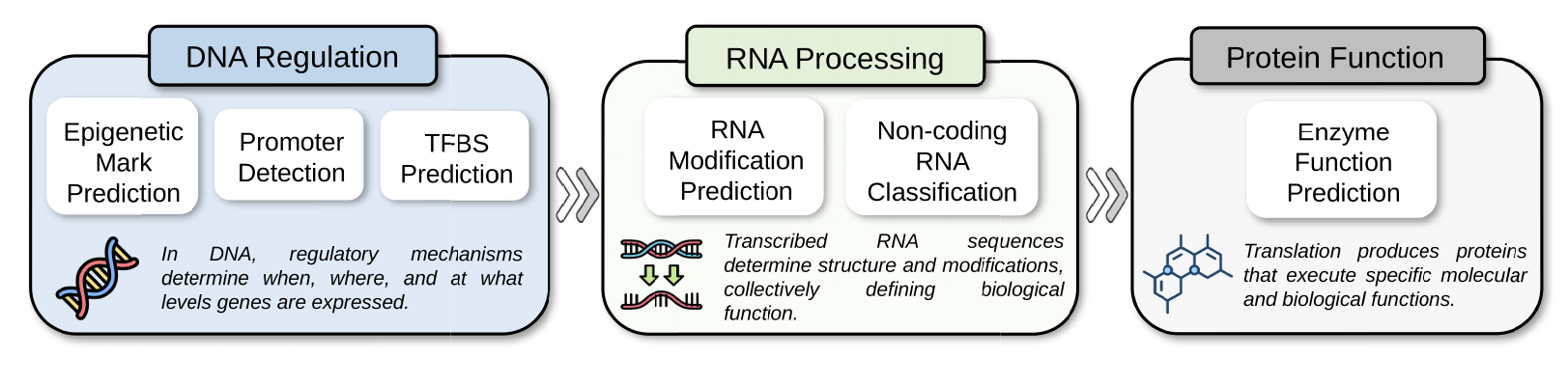}
\caption{Benchmark task distribution. OmicsBench comprises six expert-validated tasks (\(N = 1,160\)) organized into three biological categories spanning the central dogma. These tasks were selected to represent diverse reasoning granularities, ranging from local pattern recognition (e.g., Transcription Factor Binding Site) to hierarchical functional deduction (e.g., Enzyme Function), thereby ensuring that the evaluation of these fundamental capabilities generalizes to broader multi-omics reasoning challenges}
\label{fig:3}
\end{figure}

OmicsBench aligns with the central dogma of molecular biology, organizing tasks into three primary categories: DNA Regulation, RNA Processing, and Protein Function (Figure~\ref{fig:3}). This biological alignment ensures that the benchmark evaluates the model's ability across different stages of gene expression.

\FloatBarrier
\subsection{Task taxonomy}

The three aforementioned categories are further subdivided into six specific tasks (Table~\ref{tab:4}), ensuring a systematic evaluation of LLM capabilities across diverse multi-omics scenarios.

\begin{table}[!htbp]
\centering
\caption{Task taxonomy and distribution percentages. Transcription Factor Binding Site Prediction covers both human and mouse variants. Promoter Detection encompasses both general and core promoter regions.}
\label{tab:4}
\fontsize{8}{10}\selectfont
\setlength{\tabcolsep}{3pt}
\renewcommand{\arraystretch}{1.17}
\begin{adjustbox}{max width=\linewidth}
\begin{tabular}{@{}p{2.4cm}p{4.6cm}p{1.8cm}ccc@{}}
\toprule
\textbf{Category} & \textbf{Task} & \textbf{Type} & \textbf{Metric} & \textbf{N} & \textbf{\%}\\
\midrule
DNA Regulation & Epigenetic Mark Prediction & Binary & MCC & 197 & 17.0\%\\
DNA Regulation & Promoter Detection & Binary & MCC & 219 & 18.9\%\\
DNA Regulation & Transcription Factor Binding Site Prediction & Binary & MCC & 216 & 18.6\%\\
RNA Processing & RNA Modification Prediction & Multi-label & AUC & 101 & 8.7\%\\
RNA Processing & Non-coding RNA Classification & Multi-class & Acc & 215 & 18.5\%\\
Protein Function & Enzyme Function Prediction & Multi-label & F-max & 212 & 18.3\%\\
 & \textbf{Total} &  &  & \textbf{1,160} & \textbf{100.0\%}\\
\bottomrule
\end{tabular}
\end{adjustbox}
\end{table}

\textbf{DNA regulation.} LLMs analyze DNA sequences to determine the regulatory potential governing gene expression. Models must identify accessible chromatin states, recognize transcription initiation sites, and predict binding events based on sequence motifs.

\textbf{RNA processing.} LLMs interpret RNA sequences to identify post-transcriptional modifications and functional categories. This category aligns with the RNA maturation stage, where chemical marks and structural folding dictate function.

\textbf{Protein function.} LLMs infer the biochemical role of proteins by analyzing their amino acid sequences. They connect primary structure patterns to catalytic properties, corresponding to the final execution stage of the central dogma.

\subsection{Benchmark construction}

\begin{figure}[!htbp]
\centering
\includegraphics[width=\linewidth]{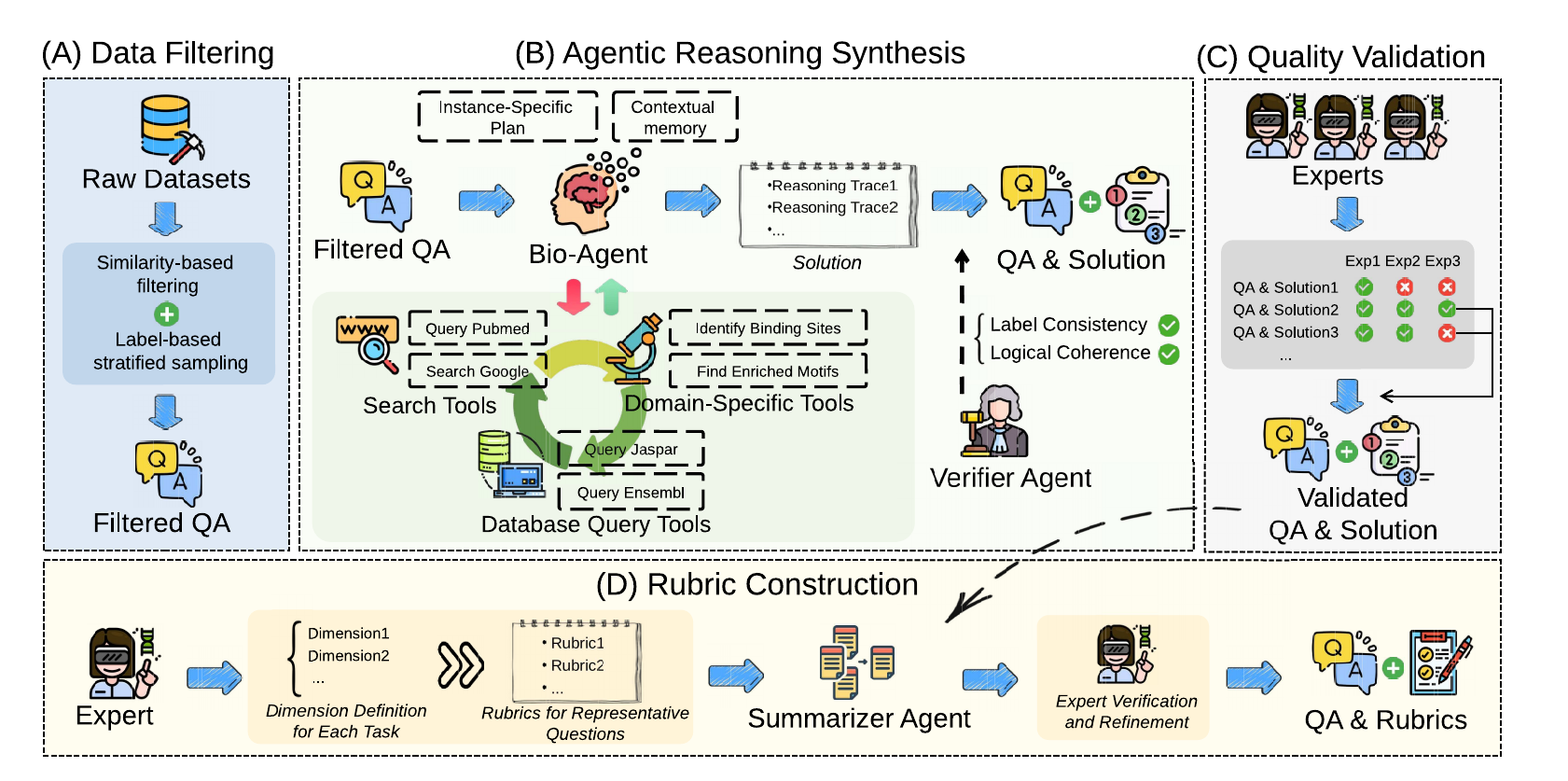}
\caption{Overview of the OmicsBench construction pipeline. The process comprises: (A) Data Filtering, where diverse sequences with balanced labels are stratified and sampled from the Biology-Instructions dataset; (B) Agentic Reasoning Synthesis, which employs a multi-agent framework consisting of a Bio-Agent, a Verifier, and a Summarizer to synthesize and self-correct reasoning traces; (C) Quality Validation, where domain experts review the content to ensure scientific rigor; and (D) Rubric Construction, where instance-specific rubrics are developed following a ``Define-Decompose-Scale-Verify'' workflow to evaluate the scientific validity of long-form responses. The pipeline ultimately yields 1,160 high-quality multi-omics reasoning tasks}
\label{fig:4}
\end{figure}

OmicsBench was developed through a structured process: data filtering, agentic synthesis, and quality validation (Figure~\ref{fig:4})~\citep{ying2025}.

\FloatBarrier
\textbf{Data filtering.} We leveraged Biology-Instructions~\citep{he2025}, a published dataset of multi-omics sequence questions containing classification labels. Using a published source ensures the validity of question formulation. The original dataset comprised 34,033 test samples across the six aforementioned tasks. To construct a high-quality benchmark, we implemented a two-step filtering process. First, to maximize sequence diversity within a manageable dataset size, we efficiently computed pairwise sequence similarities using cosine similarity and applied coreset selection~\citep{sener2018} to retain representative sequences from each cluster. Second, we performed stratified sampling to address class imbalance, ensuring uniform label distributions for binary tasks and sufficient coverage of minority classes for multi-class tasks. This process yielded 5,537 candidate questions with diverse sequences and balanced label distributions.

\textbf{Agentic data synthesis and verification.} Manually annotating biological reasoning traces is prohibitively expensive. To address this, we employed an agentic synthesis framework that automates evidence generation through tool-augmented LLMs, building upon the Biomni framework~\citep{huang2026}. The process begins with evidence generation. The core Bio-Agent (powered by Claude-Sonnet-4.5) generates preliminary reasoning traces by orchestrating domain-specific tools. For each query, the agent selects appropriate tools, such as BLAST~\citep{altschul1990} for homology search, ViennaRNA~\citep{lorenz2011} for structure prediction, JASPAR~\citep{ovekbaydar2026} for motif scanning, and the PubMed API for literature retrieval. It then synthesizes a coherent long-form response that integrates the final answer and biological justification. Subsequently, a Verifier Agent performs dual quality checks. (1) Label Consistency: It compares the Bio-Agent's predicted conclusion against the original label from the source dataset. Questions where the agent's answer diverges from the original label are discarded, as these often indicate ambiguous sequences (particularly in non-coding regions) or inherent label noise. (2) Logical Coherence: It evaluates whether the outputs from domain-specific tools logically support the final answer, checking for internal consistency and evidence sufficiency. Approximately 1,512 questions passed these automated checks.

\textbf{Expert validation.} Following the automated verification phase, a panel of three domain experts independently reviewed the candidate questions based on scientific accuracy, evidence relevance, and reasoning completeness. Experts evaluated the instances as ``accept'' or ``reject''. Questions receiving majority acceptance were retained. This rigorous process resulted in the final OmicsBench dataset of 1,160 questions (Table~\ref{tab:4}). Each entry comprises a multi-omics sequence, a specific task instruction, a ground-truth classification label, a curated long-form reasoning trace, and a set of instance-specific evaluation rubrics detailed in the Evaluation framework section.

\subsection{Evaluation framework}

OmicsBench evaluates each question-answer pair based on both the predicted label and the reasoning process. We report conventional task metrics and RR, which measures biological reasoning quality using instance-specific rubrics designed by experts.

\textbf{Metrics for prediction evaluation.} We report task-specific metrics following existing studies~\citep{he2025}. We use MCC for binary DNA tasks (EMP, Prom, TFBS). We use AUC for multi-label Mod, accuracy for multi-class ncRNA, and maximum F1-score (\(F_{\max}\)) for hierarchical EC.

\textbf{RR for reasoning evaluation.} We argue that standard evaluation paradigms, ranging from exact string matching to embedding-based similarity (\emph{e.g.}, BERTScore), are inadequate for evaluating reasoning abilities in multi-omics scenarios. These metrics often penalize correct answers expressed in different ways while rewarding hallucinated text that shares semantic proximity. Specifically, ROUGE and exact match depend on surface forms. They can penalize semantically equivalent paraphrases (\emph{e.g.}, ``TFIID binds TATA box'' \emph{vs.} ``TATA element recruits TFIID''). Embedding similarity (\emph{e.g.}, BERTScore) tends to assign high scores to linguistically fluent but factually incorrect statements (\emph{e.g.}, confusing METTL3 with METTL14 in the m6A methyltransferase complex). Furthermore, feeding model responses and reference solutions into an unconstrained ``LLM-as-a-judge'' often results in subjective and untrustworthy evaluations, especially given the LLM's limited mastery of this field.

\textbf{Instance-specific rubric construction.} To evaluate the biological reasoning capabilities of LLMs with scientific precision, we move beyond binary correctness and adopt a fine-grained, evidence-based rubric system~\citep{arora2025}. Our construction methodology bridges the gap between unstructured reasoning traces and structured evaluation criteria through a four-stage, expert-in-the-loop pipeline.

\textbf{Hierarchical rubric structure.} We formally distinguish between the task-specific dimensions and instance-specific rubrics:

\begin{itemize}
\item
  
  \textbf{Task-specific dimensions (}\(\mathcal{D}_{\text{task}}\)\textbf{):} For each biological task type (\emph{e.g.}, Prom), domain experts define a universal set of evaluation dimensions representing the theoretical ``checklist'' for that task (\emph{e.g.}, \(\mathcal{D}_{\text{Promoter}} = \{\text{Core\ Elements},\text{GC\ Content},\text{TSS\ Check},\ldots\}\text{).}\)
  
\item
  
  \textbf{Instance-specific rubrics (}\(\mathcal{R}_{q}\)\textbf{):} For a specific question \(q\), the reasoning process invokes a subset of the global dimensions to form a comprehensive evaluation standard. A single rubric entry is defined as a tuple \(r = (d,E)\), where \(d \in \mathcal{D}_{\text{task}}\) is the dimension and \(E = \{ e_{1},e_{2},\ldots\}\) is a set of atomic evidence units extracted strictly from the ground truth solution. Formally, the instance-specific rubric is defined as a set of \(k\) such entries, denoted as \(\mathcal{R}_{q} = \{ r_{1},r_{2},\ldots,r_{k}\}\), where each entry corresponds to a distinct reasoning step required to solve \(q\).
  
\end{itemize}

As shown in Figure~\ref{fig:4}(D), the rubric construction follows a ``Define-Decompose-Scale-Verify'' workflow:

\textbf{Domain-expert dimension definition.} Domain experts first establish rubric dimensions tailored to each task to define the scope of evaluation. For example, in EC tasks, dimensions include homology evidence, active site motifs, catalytic residues, substrate specificity, and reaction chemistry.

\textbf{Expert annotation via atomic decomposition.} To generate high-quality few-shot demonstrations, experts annotate 5 representative questions per task. The core innovation is Reasoning Trace Decomposition: experts break down the unstructured reference solution into atomic facts (\emph{e.g.}, specific coordinates, sequence motifs) and cluster these evidence units (\(E\)) under the relevant dimensions (\(d\)). This mapping allows for a many-to-one relationship where multiple evidence points support a single dimension.

\textbf{Automated scale-up via Summarizer Agent.} Leveraging the expert annotations as in-context learning demonstrations~\citep{dong2024}, we deploy a Summarizer Agent to process the remaining questions. The agent analyzes the reference solution, deconstructs the reasoning chain into atomic evidence, and generates structured rubric lists.

\textbf{Expert verification and refinement.} Finally, domain experts review the model-generated candidates to remove redundancy and ensure biological correctness. This quality control step ensures that the extracted evidence is factually accurate and strictly derived from the solution text without hallucination. The final dataset comprises 1,160 questions annotated with a total of 3,745 rubric dimensions (\(\sim\)3.2 dimensions per question). These rubrics are substantiated by a total of 11,392 specific evidence items, averaging approximately 3.0 atomic evidence points for each rubric dimension. This high density of evidence ensures that the evaluation captures the nuance of biological reasoning.

\textbf{Formulation of RR.} For a given instance \(q\), let \(\mathcal{R}_{q}\) denote the set of instance-specific rubrics, and let \({\widehat{y}}_{q}\) be the model output. RR measures the coverage of key reasoning steps:

\[\text{Rubric\ Recall}(q) = \frac{1}{\left| \mathcal{R}_{q} \right|}\sum_{r \in \mathcal{R}_{q}}^{}{\mathbf{1}(r\text{\ is\ satisfied\ by\ }{\widehat{y}}_{q})}\]

where \(\left| \mathcal{R}_{q} \right|\) is the total number of rubric dimensions. We define a rubric \(r = (d,E)\) as satisfied if at least one atomic evidence unit \(e \in E\) is recovered in \({\widehat{y}}_{q}\).

\textbf{Rubric judging protocol.} To operationalize RR at scale, we employ an LLM judge with carefully engineered prompts, ensuring both reproducibility and consistent adherence to evaluation standards. The judge evaluates whether each instance-specific rubric (\(r \in \mathcal{R}_{q}\)) is satisfied by the model response based on a rigorous verification process. We strictly define evidence recovery based on three criteria: (1) exact entity matching, which enforces precise alphanumeric alignment for unique identifiers (\emph{e.g.}, PDB IDs); (2) semantic equivalence, which accepts descriptive scientific facts and accommodates linguistic variations preserving biological accuracy; and (3) contextual validity, which ensures that cited evidence actively supports the conclusion rather than being misattributed or hallucinated. Furthermore, the judge is requested to verify quantitative constraints (\emph{e.g.}, recognizing that ``78\% identity'' satisfies a ``\(\geq 75\%\)'' threshold). Two design choices keep RR robust to verbosity and template gaming. First, scoring is averaged over a small number of reasoning dimensions rather than raw evidence counts, so adding more text alone cannot inflate the score. Second, the three criteria above require specific quantitative or structural details, so template-style responses that merely mention relevant terminology fail the contextual validity check. An adversarial keyword-injection test confirms this empirically: heavily jargon-injected template responses (\(\sim\)1,000 tokens each) score below 1.2\% across all six tasks.

\textbf{Inference and scoring workflow.} We evaluate each model in two stages. First, the model generates a response under standardized prompts that request both the label and the reasoning trace. Second, we extract the predicted label via pattern matching and compute the conventional task metric. We then apply rubric-based judging to compute RR and aggregate results per task.

\subsection{TA-OPD}

\begin{figure}[!htbp]
\centering
\includegraphics[width=\linewidth]{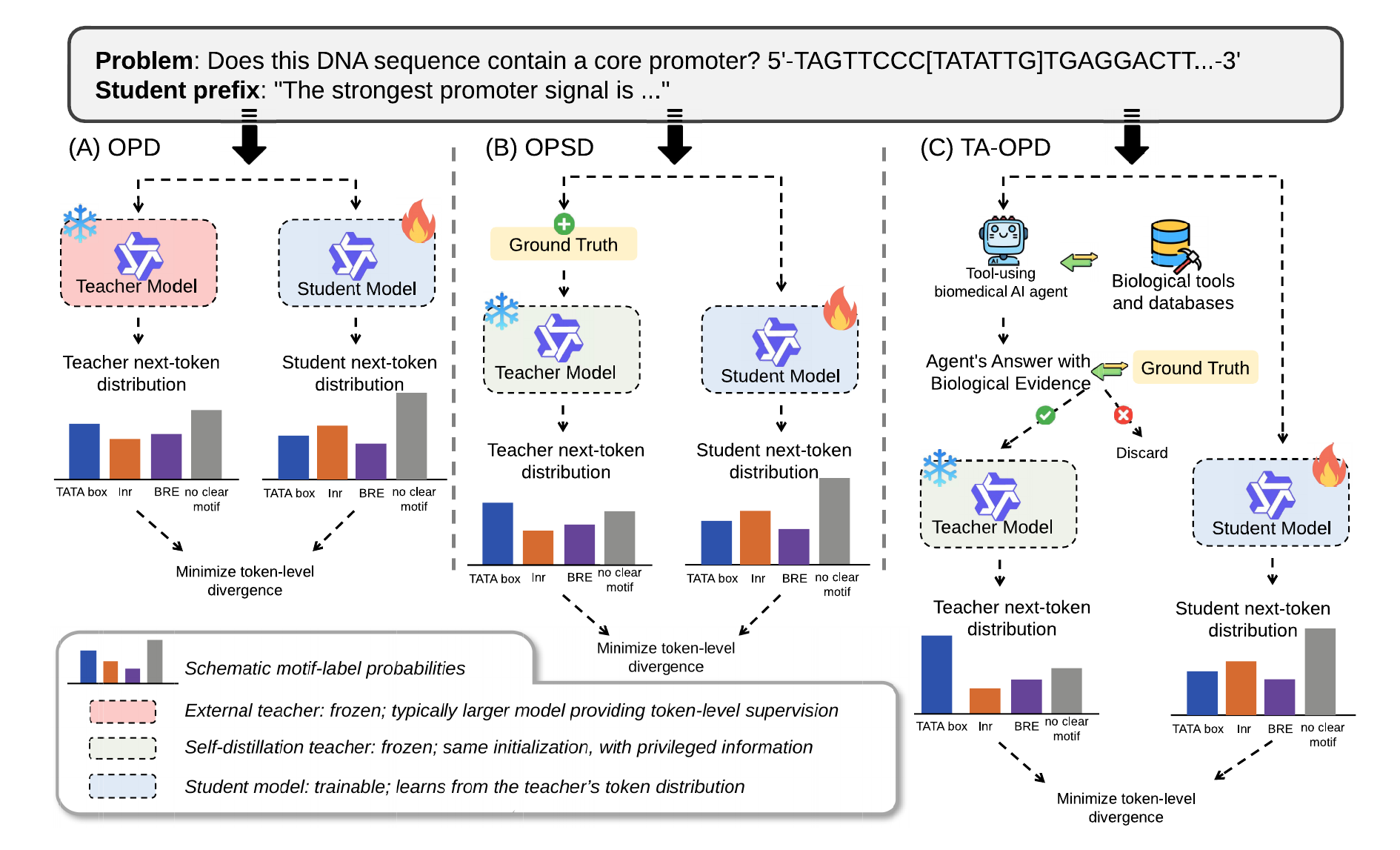}
\caption{From on-policy distillation to tool-augmented on-policy distillation. (A) OPD uses a stronger external teacher to supervise a smaller student, such as Qwen3.5-9B teaching Qwen3.5-4B. The teacher predicts the next-token distribution from the text already generated by the student, and training brings the student's distribution closer to the teacher's. (B) OPSD initializes teacher and student from the same model, such as two copies of Qwen3.5-4B, and provides ground-truth answers only to the teacher. (C) TA-OPD uses the same teacher-student initialization, with privileged information supplied by a tool-using biomedical AI agent. The agent analyzes the sequence using domain tools and databases and produces an answer with supporting evidence. An answer-label consistency gate admits an example to training only when the agent's final prediction matches the ground-truth label. The agent's answer and evidence are then provided only to the teacher. In all three panels, teachers are frozen and only student weights are updated. In the core promoter example, bars represent TATA box, initiator (Inr), TFIIB recognition element (BRE) and no clear motif. The bars are schematic and do not represent measured probabilities}
\label{fig:5}
\end{figure}

TA-OPD uses biomedical agent answers and evidence as privileged information for OPD (Figure~\ref{fig:5}). Standard OPD commonly uses a separate, stronger teacher. OPSD instead initializes teacher and student from the same model and gives ground-truth answers only to the teacher \citep{zhao2026}. TA-OPD adopts this teacher-student setup, with answers and supporting biological evidence generated by a tool-using agent implemented with Biomni \citep{huang2026}. Teacher and student are separate instances initialized from the same model weights; the teacher remains frozen throughout training. The student receives the task instruction and biological sequence and generates its own response. Both models then calculate the probability of each generated token given the task input and the earlier tokens in that response. The teacher also sees the agent's answer and evidence. We compute the distillation signal for each token as the teacher's log probability minus the student's and update only the student weights. This provides guidance throughout the student's own explanation and final prediction.

\FloatBarrier
\subsection{Training dataset construction}

The candidate training dataset was drawn from the training portion of Biology-Instructions \citep{he2025} and covered six tasks: EMP, Prom and TFBS for DNA; Mod and ncRNA for RNA; and EC for proteins. We excluded sequence-question inputs that exactly matched the 1,160 OmicsBench evaluation inputs from training. We used Biomni with DeepSeek-V4-Flash to call domain tools and generate answers with supporting biological evidence. We applied the answer-label consistency gate by comparing each final prediction with the ground-truth label from the source dataset. Because these responses served as training material rather than evaluation references, we allowed some noise in their explanations while requiring correct final predictions. We considered sample balance across tasks and used resampling with a fixed seed to reduce label imbalance within each task. The released TA-OPD-10K dataset contains 8,686 unique inputs. It comprises 10,069 training records: 2,152 EMP, 1,784 Prom, 1,670 TFBS, 1,346 Mod, 1,678 ncRNA and 1,439 EC records.

\subsection{Model training}

We applied TA-OPD to the official instruction-tuned Qwen3.5 models at five parameter sizes: 0.8B, 2B, 4B, 9B and 27B. Teacher and student shared a tokenizer so that their token probabilities could be compared directly. All models used the sampled-token k₁ log-ratio estimator with policy-gradient optimization and AdamW. We used a 3\% warm-up followed by a constant learning rate of 2 × 10⁻⁶, with a weight decay of 0.01. Across training runs, prompts were limited to 8,192 tokens and responses to at most 16,384 tokens. Training used up to eight NVIDIA H200 graphics processing units (GPUs), each with 140 GB of memory.

\subsection{Model evaluation}

We evaluated unadapted and TA-OPD models on the same 1,160 OmicsBench questions, using the same task prompts and label-extraction procedures. For each model, we generated responses with a temperature of 1.0, a maximum output length of 16,384 tokens and five random seeds (0 to 4). Models received neither privileged information nor biological tools during evaluation. We assessed biological evidence using the instance-specific OmicsBench rubrics and the same DeepSeek-V3.2 judge protocol for every model. Within each seed, Avg Recall was the mean RR across the six tasks, with equal weight for each task. For each metric, including Avg Recall, we removed the highest and lowest of the five seed-specific values and reported the mean and sample standard deviation of the remaining three.

\subsection{Data and model availability}

\begingroup
\hypersetup{urlcolor=black}
\urlstyle{same}
OmicsBench is available at \url{https://github.com/SeedScientist/OmicsBench}. The TA-OPD-10K training dataset, comprising 10,069 records, is available at \url{https://huggingface.co/datasets/yj12869741/TA-OPD-10K}. The TA-OPD-27B model weights are available at \url{https://huggingface.co/yj12869741/TA-OPD-27B}.
\par
\endgroup

\FloatBarrier
\begingroup
\small
\setlength{\bibhang}{1.5em}
\setlength{\bibsep}{3pt}

\endgroup

\begin{thebibliography}{35}
\bibitem[Agarwal et al.(2024)]{agarwal2024}
Agarwal, R., Vieillard, N., Zhou, Y., Stanczyk, P., Ramos Garea, S., Geist, M., and Bachem, O. (2024). On-policy distillation of language models: Learning from self-generated mistakes. In International Conference on Learning Representations, pp. 21246--21263.
\bibitem[Altschul et al.(1990)]{altschul1990}
Altschul, S.F., Gish, W., Miller, W., Myers, E.W., and Lipman, D.J. (1990). \href{https://doi.org/10.1016/S0022-2836(05)80360-2}{Basic local alignment search tool}. J Mol Biol 215, 403--410.
\bibitem[Arora et al.(2026)]{arora2026}
Arora, R., Chen, L.T., Du, M., Marks, D.S., and Church, G.M. (2026). PG-LLM: Benchmarking general-purpose language models for protein variant ranking. bioRxiv preprint. \url{https://doi.org/10.64898/2026.07.27.741045}.
\bibitem[Arora et al.(2025)]{arora2025}
Arora, R.K., Wei, J., Hicks, R.S., Bowman, P., Quiñonero-Candela, J., Tsimpourlas, F., Sharman, M., Shah, M., Vallone, A., Beutel, A., et al. (2025). HealthBench: Evaluating large language models towards improved human health. arXiv preprint arXiv:2505.08775.
\bibitem[Boiko et al.(2023)]{boiko2023}
Boiko, D.A., MacKnight, R., Kline, B., and Gomes, G. (2023). Autonomous chemical research with large language models. Nature 624, 570--578.
\bibitem[Chen et al.(2022)]{chen2022}
Chen, J., Hu, Z., Sun, S., Tan, Q., Wang, Y., Yu, Q., Zong, L., Hong, L., Xiao, J., Shen, T., et al. (2022). Interpretable RNA foundation model from unannotated data for highly accurate RNA structure and function predictions. arXiv preprint arXiv:2204.00300.
\bibitem[Crick(1970)]{crick1970}
Crick, F. (1970). Central dogma of molecular biology. Nature 227, 561--563.
\bibitem[de Almeida et al.(2025)]{dealmeida2025}
de Almeida, B.P., Richard, G., Dalla-Torre, H., Blum, C., Hexemer, L., Pandey, P., Laurent, S., Rajesh, C., Lopez, M., Laterre, A., et al. (2025). \href{https://doi.org/10.1038/s42256-025-01047-1}{A multimodal conversational agent for DNA, RNA and protein tasks}. Nat Mach Intell 7, 928--941.
\bibitem[Dong et al.(2024)]{dong2024}
Dong, Q., Li, L., Dai, D., Zheng, C., Ma, J., Li, R., Xia, H., Xu, J., Wu, Z., Chang, B., et al. (2024). \href{https://doi.org/10.18653/v1/2024.emnlp-main.64}{A survey on in-context learning}. In Proceedings of the 2024 Conference on Empirical Methods in Natural Language Processing, pp. 1107--1128.
\bibitem[Geirhos et al.(2020)]{geirhos2020}
Geirhos, R., Jacobsen, J.-H., Michaelis, C., Zemel, R., Brendel, W., Bethge, M., and Wichmann, F.A. (2020). \href{https://doi.org/10.1038/s42256-020-00257-z}{Shortcut learning in deep neural networks}. Nat Mach Intell 2, 665--673.
\bibitem[Ghareeb et al.(2026)]{ghareeb2026}
Ghareeb, A.E., Chang, B., Mitchener, L., Yiu, A., Szostkiewicz, C.J., Shved, D., Gyimesi, G.J., Laurent, J.M., Wright, S.M., Razzak, M.T., et al. (2026). A multi-agent system for automating scientific discovery. Nature 655, 497--505.
\bibitem[Hasin et al.(2017)]{hasin2017}
Hasin, Y., Seldin, M., and Lusis, A. (2017). Multi-omics approaches to disease. Genome Biol 18, 83.
\bibitem[He et al.(2025)]{he2025}
He, H., Ren, Y., Tang, Y., Xu, Z., Li, J., Yang, M., Zhang, D., Dong, Y., Chen, T., Zhang, S., et al. (2025). \href{https://doi.org/10.18653/v1/2025.findings-emnlp.978}{Biology-Instructions: A dataset and benchmark for multi-omics sequence understanding capability of large language models}. In Findings of the Association for Computational Linguistics: EMNLP 2025, pp. 17984--18016.
\bibitem[Huang et al.(2026)]{huang2026}
Huang, K., Zhang, S., Wang, H., Qu, Y., Lu, Y., Li, R., Roohani, Y., Qiu, L., Cao, S., Li, G., et al. (2026). \href{https://doi.org/10.1126/science.adz4351}{Autonomous biomedical research with an artificial intelligence agent}. Science 393, eadz4351.
\bibitem[International Human Genome Sequencing Consortium(2004)]{ihgsc2004}
International Human Genome Sequencing Consortium. (2004). Finishing the euchromatic sequence of the human genome. Nature 431, 931--945.
\bibitem[Jiang et al.(2025)]{jiang2025}
Jiang, W., Ye, W., Tan, X., and Bao, Y.-J. (2025). \href{https://doi.org/10.1186/s13040-025-00442-z}{Network-based multi-omics integrative analysis methods in drug discovery: A systematic review}. BioData Min 18, 27.
\bibitem[Jin et al.(2024)]{jin2024}
Jin, Q., Yang, Y., Chen, Q., and Lu, Z. (2024). \href{https://doi.org/10.1093/bioinformatics/btae075}{GeneGPT: Augmenting large language models with domain tools for improved access to biomedical information}. Bioinformatics 40, btae075.
\bibitem[Laurent et al.(2024)]{laurent2024}
Laurent, J.M., Janizek, J.D., Ruzo, M., Hinks, M.M., Hammerling, M.J., Narayanan, S., Ponnapati, M., White, A.D., and Rodriques, S.G. (2024). LAB-Bench: Measuring capabilities of language models for biology research. arXiv preprint arXiv:2407.10362.
\bibitem[Lin et al.(2023)]{lin2023}
Lin, Z., Akin, H., Rao, R., Hie, B., Zhu, Z., Lu, W., Smetanin, N., Verkuil, R., Kabeli, O., Shmueli, Y., et al. (2023). \href{https://doi.org/10.1126/science.ade2574}{Evolutionary-scale prediction of atomic-level protein structure with a language model}. Science 379, 1123--1130.
\bibitem[Lorenz et al.(2011)]{lorenz2011}
Lorenz, R., Bernhart, S.H., Höner zu Siederdissen, C., Tafer, H., Flamm, C., Stadler, P.F., and Hofacker, I.L. (2011). \href{https://doi.org/10.1186/1748-7188-6-26}{ViennaRNA package 2.0}. Algorithms Mol Biol 6, 26.
\bibitem[Luo et al.(2026)]{luo2026}
Luo, F., Chuang, Y.-N., Wang, G., Xu, Z., Han, X., Zhang, T., and Braverman, V. (2026). Demystifying OPD: Length inflation and stabilization strategies for large language models. arXiv preprint arXiv:2604.08527.
\bibitem[Nguyen et al.(2024)]{nguyen2024}
Nguyen, E., Poli, M., Durrant, M.G., Kang, B., Katrekar, D., Li, D.B., Bartie, L.J., Thomas, A.W., King, S.H., Brixi, G., et al. (2024). \href{https://doi.org/10.1126/science.ado9336}{Sequence modeling and design from molecular to genome scale with Evo}. Science 386, eado9336.
\bibitem[Ovek Baydar et al.(2026)]{ovekbaydar2026}
Ovek Baydar, D., Rauluseviciute, I., Aronsen, D.R., Blanc-Mathieu, R., Bonthuis, I., de Beukelaer, H., Ferenc, K., Jégou, A., Kumar, V., Lemma, R.B., et al. (2026). \href{https://doi.org/10.1093/nar/gkaf1209}{JASPAR 2026: Expansion of transcription factor binding profiles and integration of deep learning models}. Nucleic Acids Res 54, D184--D193.
\bibitem[Schmidhuber(2007)]{schmidhuber2007}
Schmidhuber, J. (2007). Gödel machines: Fully self-referential optimal universal self-improvers. In Artificial General Intelligence, B. Goertzel and C. Pennachin, eds. (Berlin, Heidelberg: Springer), pp. 199--226.
\bibitem[Penadés et al.(2025)]{penades2025}
Penadés, J.R., Gottweis, J., He, L., Patkowski, J.B., Daryin, A., Weng, W.-H., Tu, T., Palepu, A., Myaskovsky, A., Pawlosky, A., et al. (2025). AI mirrors experimental science to uncover a mechanism of gene transfer crucial to bacterial evolution. Cell 188, 6654--6665.e2.
\bibitem[Sener and Savarese(2018)]{sener2018}
Sener, O., and Savarese, S. (2018). Active learning for convolutional neural networks: A core-set approach. In International Conference on Learning Representations.
\bibitem[Shang et al.(2025)]{shang2025}
Shang, X., Liao, X., Ji, Z., and Hou, W. (2025). \href{https://doi.org/10.1093/bib/bbaf492}{Benchmarking large language models for genomic knowledge with GeneTuring}. Brief Bioinform 26, bbaf492.
\bibitem[Shenfeld et al.(2026)]{shenfeld2026}
Shenfeld, I., Damani, M., Hübotter, J., and Agrawal, P. (2026). Self-distillation enables continual learning. In Proceedings of the 43rd International Conference on Machine Learning.
\bibitem[Singhal et al.(2023)]{singhal2023}
Singhal, K., Azizi, S., Tu, T., Mahdavi, S.S., Wei, J., Chung, H.W., Scales, N., Tanwani, A., Cole-Lewis, H., Pfohl, S., et al. (2023). Large language models encode clinical knowledge. Nature 620, 172--180.
\bibitem[Swanson et al.(2025)]{swanson2025}
Swanson, K., Wu, W., Bulaong, N.L., Pak, J.E., and Zou, J. (2025). \href{https://doi.org/10.1038/s41586-025-09442-9}{The virtual lab of AI agents designs new SARS-CoV-2 nanobodies}. Nature 646, 716--723.
\bibitem[Wang et al.(2025)]{wang2025}
Wang, Z., Jin, Q., Wei, C.-H., Tian, S., Lai, P.-T., Zhu, Q., Day, C.-P., Ross, C., Leaman, R., and Lu, Z. (2025). GeneAgent: self-verification language agent for gene-set analysis using domain databases. Nat Methods 22, 1677--1685.
\bibitem[Xia et al.(2025)]{xia2025}
Xia, Y., Jin, P., Xie, S., He, L., Cao, C., Luo, R., Liu, G., Wang, Y., Liu, Z., Chen, Y.-J., et al. (2025). Nature language model: Deciphering the language of nature for scientific discovery. arXiv preprint arXiv:2502.07527.
\bibitem[Ying et al.(2025)]{ying2025}
Ying, J., Chen, Z., Wang, Z., Jiang, W., Wang, C., Yuan, Z., Su, H., Kong, H., Yang, F., and Dong, N. (2025). \href{https://doi.org/10.18653/v1/2025.acl-long.1516}{SeedBench: A multi-task benchmark for evaluating large language models in seed science}. In Proceedings of the 63rd Annual Meeting of the Association for Computational Linguistics, pp. 31395--31449.
\bibitem[Yu et al.(2026)]{yu2026}
Yu, X., Li, G., Si, Q., Zhang, G., Xu, Y., Wang, C., Dong, S., Tuo, K., Zeng, X., Feng, K., et al. (2026). DOPD: Dual on-policy distillation. arXiv preprint arXiv:2606.30626.
\bibitem[Zhao et al.(2026)]{zhao2026}
Zhao, S., Xie, Z., Liu, M., Huang, J., Pang, G., Chen, F., and Grover, A. (2026). Self-distilled reasoner: On-policy self-distillation for large language models. In Proceedings of the 43rd International Conference on Machine Learning.
\end{thebibliography}
\end{document}